\documentclass[conference]{IEEEtran}
\IEEEoverridecommandlockouts

\usepackage{cite}
\usepackage{amsmath,amssymb,amsfonts}
\usepackage{algorithmic}
\usepackage{graphicx}
\usepackage{textcomp}
\usepackage{xcolor}
\def\BibTeX{{\rm B\kern-.05em{\sc i\kern-.025em b}\kern-.08em
    T\kern-.1667em\lower.7ex\hbox{E}\kern-.125emX}}

\usepackage[table]{xcolor}

\usepackage{booktabs}
\usepackage[caption=false,font=footnotesize]{subfig}

\usepackage{url}
\usepackage{multirow} 
\usepackage{xspace}
\usepackage{bm}
\usepackage{pifont}
\usepackage{makecell}
\usepackage{amsthm}
\usepackage{amssymb}
\newcommand{\ourmethod}{\textsc{SIM}\xspace}

\begin{document}
\title{\ourmethod: Subspace Interaction-based Method for Token-Level Text Anomaly Detection}

\author{
\IEEEauthorblockN{
Kehan Yan$^{1}$, 
Yue Tan$^{2}$, 
Qingfeng Chen$^{\dagger 1}$, 
Shiyuan Li$^{2}$, 
Yu Zheng$^{2}$, 
and Yixin Liu$^{\dagger 2}$%
\thanks{$^{\dagger}$ Qingfeng Chen and Yixin Liu are the corresponding authors.}}
\textit{$^1$ Guangxi University, Guangxi, China}, 
\textit{$^2$ Griffith University, Queensland, Australia} \\
\IEEEauthorblockA{
2413301048@st.gxu.edu.cn, 
\{yue.tan, shiyuan.li, yu.zheng, yixin.liu\}@griffith.edu.au, 
qingfeng@gxu.edu.cn
}
}

\maketitle

\begin{abstract}
Token-level text anomaly detection, as an emerging trend of text anomaly detection, moves beyond coarse-grained document-level detection by localizing anomalous tokens within text. By providing fine-grained abnormality prediction, token-level text anomaly detection plays a critical role in various real-world applications, such as spam filtering and fake news detection. However, existing methods still rely on the global distance calculation for scoring, during which the local anomaly signals are severely diluted by numerous redundant normal feature dimensions. Moreover, pre-trained language models used in these methods inevitably smooth out surface anomalies, further limiting their effectiveness in token-level anomaly detection. To address these limitations, we propose a Subspace Interaction-based Method (SIM for short) for token-level text anomaly detection. To prevent local signal dilution, SIM adopts a subspace interaction-based anomaly detector, which decouples high-dimensional token embeddings into multiple low-dimensional ones, amplifying localized anomaly signals hidden within specific dimensions. To counteract the over-smoothing effect, we design a hard pseudo-anomaly generation module to construct pseudo-anomalous tokens, simulating the subtle anomalies obscured by semantic smoothing. Also, a probabilistic boundary loss is developed to standardize anomaly scores into statistical distances, effectively enforcing anomalous instances to deviate significantly from the normal distribution center. Extensive experiments on multiple benchmark datasets verify the effectiveness of SIM and demonstrate its remarkable efficiency, robustness, and interpretability. The source code is available at: \textcolor{blue}{\url{https://github.com/yankehan/SIM-TAD}}.
\end{abstract}


\section{Introduction}
Anomaly detection is a foundational research problem that plays a critical role in numerous practical application scenarios~\cite{pang2021deep,chalapathy2019deep,han2022adbench,zheng2026unsupervised}. For example, in tabular~\cite{yin2024mcm,ye2025drl} and graph data~\cite{pan2023prem,zhao2025freegad}, anomaly detection is commonly utilized to identify financial fraud or malicious nodes within social networks. However, textual data possesses highly unstructured, discrete, and complex semantic characteristics, limiting early progress in anomaly detection specifically targeting textual data~\cite{ruff2019self,manolache2021date,pan2026camera}. These characteristics have motivated recent research on text anomaly detection, which aims to identify text instances that deviate from normal semantic or syntactic patterns~\cite{cao2025tad,cao2025text}. Due to the ubiquity of textual data, text anomaly detection has become increasingly important in scenarios such as spam filtering, fake news detection, and machine-generated content identification~\cite{qian2026dynhd,chen2025multi}.

In recent years, the rapid advancement of pre-trained language models (PLMs)~\cite{devlin2019bert,liu2019roberta} has provided text anomaly detection with powerful contextual representations that encode rich semantic and syntactic information. 
By combining these high-quality text embeddings with mainstream anomaly detectors, existing methods have achieved promising performance in solving text anomaly detection problems~\cite{li2024nlp}. However, most existing studies are restricted to \textbf{document-level anomaly detection} tasks that assign a single anomaly score to each document to indicate its overall abnormality~\cite{ruff2019self,manolache2021date,das2023few}. This coarse-grained detection makes it difficult for users to understand the basis for anomaly determination and prevents the precise localization of problematic segments. In fact, practical applications are in urgent demand for fine-grained anomaly localization~\cite{cao2026towards}. For example, practitioners may need to locate anomalous instructions that cause program crashes in massive website backend logs, while security analysts need to identify deceptive illegal links in phishing emails. In these scenarios, fine-grained anomaly localization not only facilitates rapid troubleshooting but also significantly enhances the transparency and trustworthiness of the detection system.

\begin{figure}[t]
\centering
\subfloat[SMS\_Spam]{
\includegraphics[width=0.48\columnwidth]{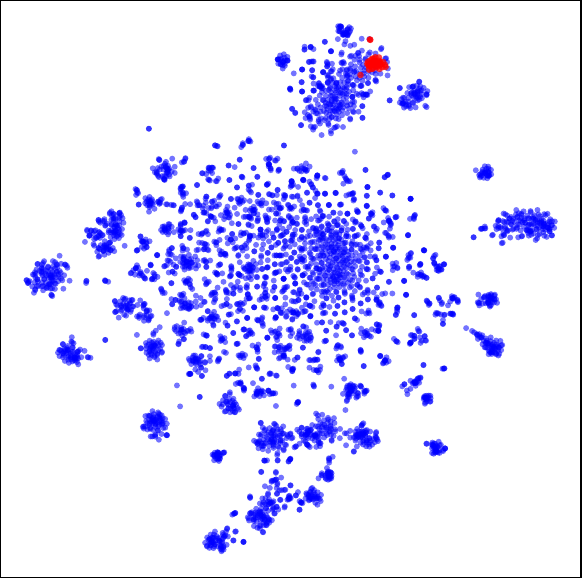}}
\label{fig:intro_tsne_smsspam}
\hfill
\subfloat[Grammar]{
\includegraphics[width=0.48\columnwidth]{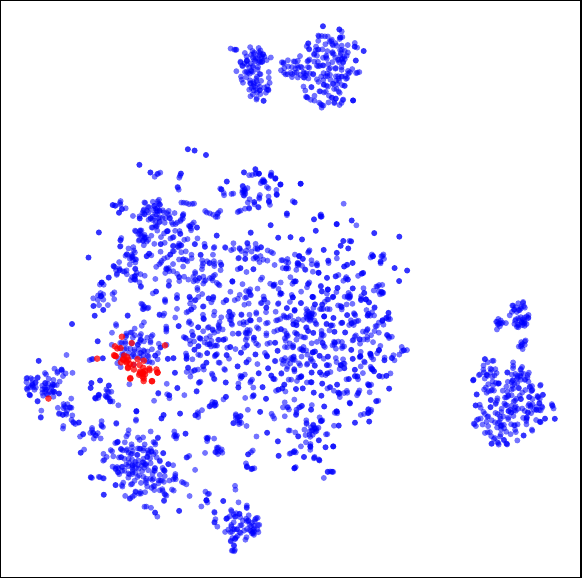}}
\label{fig:intro_tsne_grammar}
\caption{t-SNE visualization of token embeddings on two datasets corresponding to different anomaly types: SMS\_Spam (SMS gibberish corruption) and Grammar (grammatical anomalies).}
\label{fig:intro_tsne}
\end{figure}

To achieve fine-grained text anomaly detection, Cao et al.~\cite{cao2026towards} proposed the first token-level text anomaly detection framework, named TokenCore. This method utilizes PLMs to encode each token in a document into a high-dimensional numerical embedding, and then derives anomaly scores by calculating the nearest neighbor distance between the test token and the set of normal tokens. Because PLMs can capture rich semantic and syntactic features, anomalous tokens that deviate from normal patterns exhibit a significant distance shift from the distribution of normal tokens within the embedding space. Consequently, these anomalous tokens are assigned higher anomaly scores. Overall, TokenCore opens up a new direction for text anomaly detection by extending anomaly scoring from the document level to the token level.

\begin{figure}[t]
    \centering
    \includegraphics[width=0.48\textwidth]{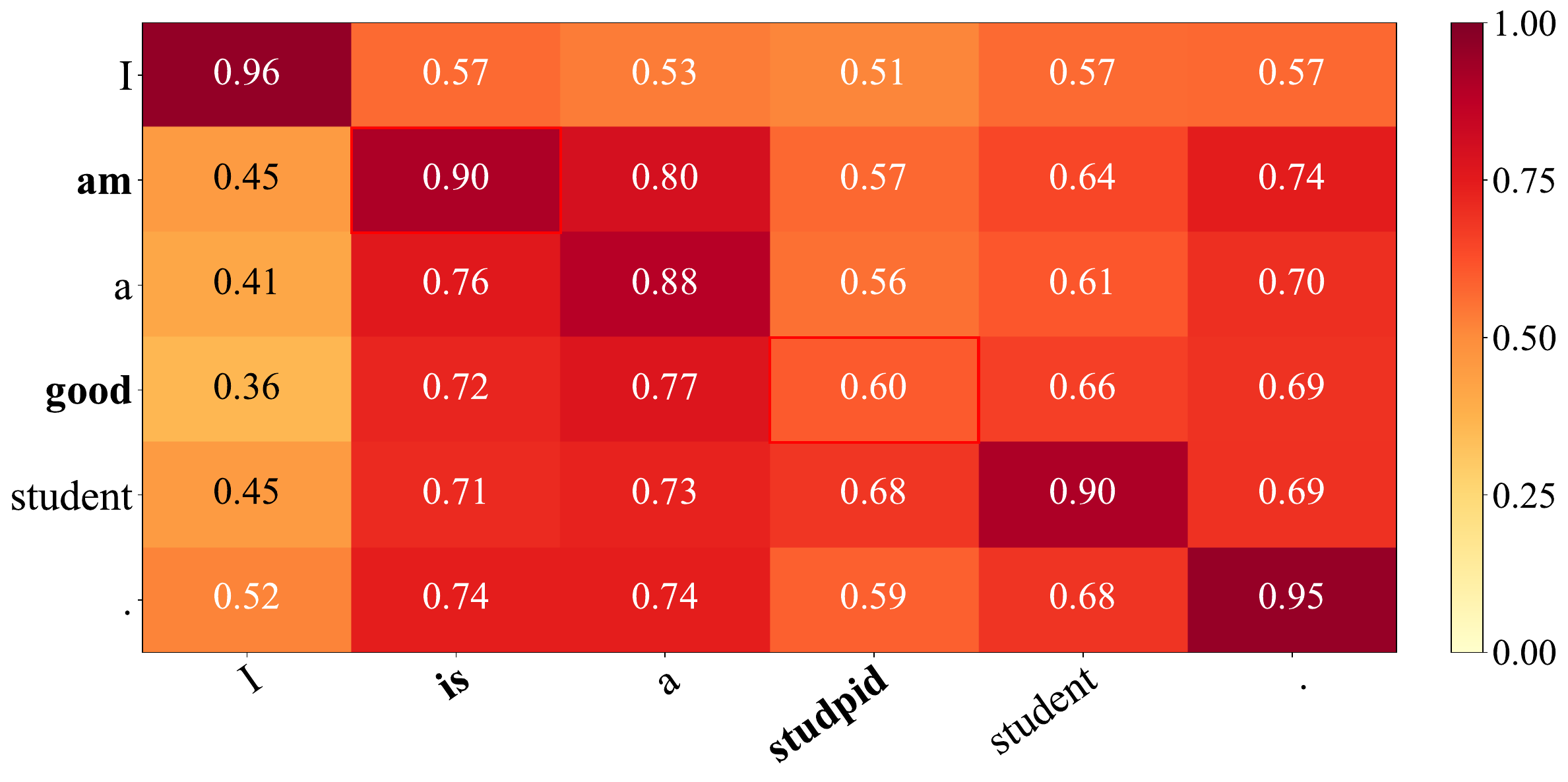}
    \caption{Visualization of BERT embedding similarities. Grammatical variants (``am'' vs.~``is'') show high latent similarity, whereas semantically distinct tokens (``good'' vs.~``stupid'') show lower similarity scores.}
    \label{fig:intro_heatmap}
\end{figure}

Although TokenCore~\cite{cao2026towards} shows promising potential in token-level anomaly detection, it highly relies on a distance-based scoring mechanism, which directly relies on the distance of embedding vectors in the feature space, inevitably introducing two inherent limitations. 
\textit{\textbf{Limitation 1:} Local anomaly signal dilution.} 
While a small number of obvious anomalies, such as tokens with extreme contextual mismatches or strong negative sentiments, can be easily detected in the global embedding space, most subtle anomalies, such as text corruption, are only reflected in a few specific embedding dimensions, i.e., subspaces. 
During global distance calculation, these localized anomaly signals are severely diluted by numerous redundant normal feature dimensions~\cite{tu2024weighted}. Consequently, as illustrated in Figure~\ref{fig:intro_tsne}, anomalous tokens remain close to normal-token clusters in the feature space, making them difficult to distinguish through global distance-based scoring. 
\textit{\textbf{Limitation 2:} Over-smoothing effect.} Because PLMs such as BERT were originally designed to optimize semantic similarity rather than anomaly sensitivity~\cite{devlin2019bert,shi2022revisiting}, their representations inevitably smooth out surface anomalies, such as grammatical errors. Taking the representation-level similarity in Figure~\ref{fig:intro_heatmap} as an example, surface anomalous tokens (e.g., \texttt{is}) exhibit high similarity to normal tokens (e.g., \texttt{am}) in the latent space and are difficult to distinguish, leading to weak anomaly separability for such types of anomalous tokens. 
Motivated by these limitations, a natural question arises: \textbf{\textit{Can we develop a token-level text anomaly detection framework capable of capturing local anomaly signals while counteracting the over-smoothing effect inherent in PLMs?}}

To answer this question, we propose \ourmethod, a \textbf{S}ubspace \textbf{I}nteraction-based \textbf{M}ethod 
for token-level text anomaly detection. 
To address \textit{\textbf{Limitation 1}}, we design a \textit{subspace interaction-based anomaly detector}, which decouples high-dimensional token embeddings into multiple low-dimensional feature subspaces and introduces a cross-subspace self-attention mechanism to dynamically capture interactions between subspaces. This subspace interaction mechanism endows the model with the ability to amplify local anomaly signals hidden within specific dimensions, enhancing its sensitivity to subtle token-level anomalies. 
To handle \textit{\textbf{Limitation 2}}, we train the anomaly detector with two complementary strategies, namely \textit{hard pseudo-anomaly generation} and \textit{probabilistic boundary loss}, improving its sensitivity to surface anomalies that are smoothed out in the representation space. 
Specifically, given that smoothed surface anomalies are highly similar to normal tokens in the representation space, the hard pseudo-anomaly generation introduces a distance perturbation mechanism to artificially construct pseudo-anomalous tokens. 
These tokens are highly similar to normal tokens in the representation space, yet they lack meaningful semantic and syntactic structures; therefore, they can serve as suitable pseudo-anomalies.
Training on these hard pseudo-anomalies discourages the detector from relying solely on semantic similarity and encourages it to capture subtle anomaly signals beyond global representation distance. 
Furthermore, we design a {probabilistic boundary loss} to guide anomaly score prediction. This loss utilizes the mean and variance of the anomaly scores of normal data to construct a statistical boundary, thereby enforcing that the scores of anomalous samples significantly deviate from normal instances in a statistical sense.

In summary, the contributions of this paper are threefold:
\begin{itemize}
    \item \textbf{New Perspective:} Going beyond the traditional perspective of global token embeddings, we are the first to explore the utilization of token subspace interaction information to achieve token-level text anomaly detection.
    \item \textbf{Novel Method:} We propose \ourmethod, a lightweight model using a subspace interaction-based anomaly detector to capture subspace anomaly signals. Moreover, we introduce a hard pseudo-anomaly generation module to overcome the over-smoothing effect, and design a probabilistic boundary loss to guide the learning of anomaly scores.
    \item \textbf{Extensive Experiments:} Extensive experiments on three real-world datasets demonstrate that \ourmethod achieves superior anomaly detection performance at both token and document levels compared to state-of-the-art approaches, while delivering remarkable efficiency, robustness, and interpretability.
\end{itemize}

\section{Related Work}
\subsection{Anomaly Detection}
Anomaly detection aims to identify samples that deviate from standard data distributions~\cite{pang2021deep,cao2025anomaly,chalapathy2019deep}. Early research relied on explicit assumptions about data-space structures to formulate interpretable detection criteria. Specifically, LOF~\cite{breunig2000lof} assumes local consistency, identifying anomalies through local density deviations. iForest~\cite{liu2008isolation} assumes anomalies are easier to isolate in the feature space and measures abnormality through random partitioning. ECOD~\cite{li2022ecod} focuses on overall distribution geometry and quantifies anomalies by modeling tail events. With the advancement of representation learning, the paradigm has shifted toward learning normative representation structures. For instance, AutoEncoder~\cite{zhou2017anomaly} uses reconstruction errors as anomaly signals, assuming normal data can be reconstructed more accurately. DeepSVDD~\cite{ruff2018deep} learns a compact representation of normal data and identifies samples that deviate significantly from it. LUNAR~\cite{goodge2022lunar} leverages local neighborhood information to identify complex anomalous patterns.

Despite being widely adopted and strong baselines in domains such as tabular~\cite{li2026towards} and graph data~\cite{zhao2025freegad,zhao2026fedcigar,liu2026few,li2026ofa}, these methods primarily evaluate anomalies in the global feature space. Consequently, when anomalous patterns are sparse and localized, localized anomaly signals are severely diluted by redundant normal feature dimensions~\cite{li2026towards2}. This motivates fine-grained anomaly detectors that capture localized anomaly signals for improved anomaly detection performance.

\subsection{Document-Level Text Anomaly Detection}
Document-level Text Anomaly Detection aims to identify text instances that deviate from normal semantic or syntactic patterns~\cite{cao2025tad,cao2025text,li2024nlp}. Existing methods can be primarily categorized into two technical routes: end-to-end methods and two-stage methods. End-to-end methods directly take raw text as input and output anomaly scores. For instance, CVDD~\cite{ruff2019self} introduces multiple learnable contextual prototype vectors and quantifies abnormality based on the distance between sample features and these prototypes. DATE~\cite{manolache2021date} applies substitution perturbations to text and constructs self-supervised tasks to model normal text patterns and identify anomalies. FATE~\cite{das2023few} introduces a deviation learning mechanism that uses a minimal amount of labeled anomalous samples to optimize anomaly scores. In contrast, two-stage methods~\cite{liu2026beyond} first use PLMs~\cite{devlin2019bert,liu2019roberta} to encode raw text into dense embedding vectors, then apply traditional anomaly detection algorithms for continuous vector spaces (e.g., LOF~\cite{breunig2000lof} and iForest~\cite{liu2008isolation}) for outlier detection.

Although both categories have made significant progress, they only predict anomaly scores for the entire text and cannot localize specific anomalous tokens. However, in practical tasks such as grammatical error correction, merely identifying an entire sentence as anomalous is meaningless. Therefore, it is essential to shift from coarse-grained document-level discrimination to fine-grained token-level anomaly detection.

\subsection{Token-Level Text Anomaly Detection}
Token-level text anomaly detection aims to assess the global abnormality of a document while precisely localizing anomalous tokens. This task differs fundamentally from traditional supervised token-level tasks (e.g., spell checking, grammatical error detection, or Named Entity Recognition~\cite{li2020survey}). First, it operates under a one-class setting where only normal documents are available during training to learn normal token distributions, lacking supervisory signals for actual anomalies. Second, rather than focusing on predefined anomaly types, it identifies diverse and unknown anomaly variants, increasing its complexity. As a pioneering attempt, Cao et al.~\cite{cao2026towards} proposed the representative baseline TokenCore. Specifically, TokenCore maps individual tokens into high-dimensional numerical representations via PLMs, then quantifies abnormality using the nearest neighbor distance between a test token and a set of normal tokens. This paradigm establishes an effective baseline for token-level text anomaly detection.

While TokenCore achieves promising performance, its reliance on spatial distance measurements in the global feature space exposes two critical limitations~\cite{liu2026rethinking}. First, weak anomaly signals in specific subspaces are severely diluted by redundant normal dimensions during global distance computation. Second, the intrinsic over-smoothing effect of PLMs causes surface-level anomalies to map closely to normal tokens in the latent space, making them nearly indistinguishable. To overcome these limitations, we propose \ourmethod, a Subspace Interaction-based Method for token-level text anomaly detection. By integrating a subspace interaction-based anomaly detector and a hard pseudo-anomaly generation, \ourmethod resolves local signal dilution and representation over-smoothing. Furthermore, we introduce a probabilistic boundary loss to guide the model in acquiring reliable anomaly scores.

\section{Preliminaries}
\noindent\textbf{Pre-trained Language Model (PLM)-Generated Embeddings.} 
To perform fine-grained text anomaly detection, we utilize a mainstream PLM (such as BERT~\cite{devlin2019bert}) to generate token-level embeddings for subsequent anomaly detection. Concretely, let $\mathcal{D} = \{\mathcal{X}_i\}_{i=1}^N$ be a text dataset consisting of $N$ documents. Each document $\mathcal{X}_i$ comprises a variable-length token sequence $\mathcal{X}_i = \{x_{i,1}, x_{i,2}, \dots, x_{i,T_i}\}$. Given a PLM $f(\cdot)$, each document $\mathcal{X}_i$ is mapped to a sequence of token embeddings:
\begin{equation}
    \mathbf{Z}_i = f(\mathcal{X}_i) = \{\mathbf{z}_{i,1}, \mathbf{z}_{i,2}, \dots, \mathbf{z}_{i,T_i}\}.
\end{equation}

\noindent\textbf{Token-Level Text Anomaly Detection.} 
The goal of token-level text anomaly detection is to learn a token-level anomaly scoring function $s(\cdot)$ that computes an anomaly score $s_{i, t}$ for each token embedding $\mathbf{z}_{i,t}$:
\begin{equation}
    s_{i, t} = s(\mathbf{z}_{i,t}),
\end{equation}
where a higher score indicates a greater degree of anomaly for the corresponding token.

\noindent\textbf{Document-Level Text Anomaly Detection.} 
The goal of document-level text anomaly detection is to learn a document-level anomaly score $S_i$ for each document $\mathcal{X}_i$. Leveraging the obtained token-level anomaly scores, we can derive the document-level anomaly score $S_i$ by applying an aggregation strategy $\text{Aggregate}(\cdot)$ over the individual token-level scores within $X_i$:
\begin{equation}
    S_i = \text{Aggregate}(\{s_{i, t}\}_{t=1}^{T_i}).
\end{equation}
Crucially, besides providing a document-level anomaly score, these token-level anomaly scores inherently offer fine-grained interpretability by identifying the exact tokens that trigger document-level anomalies and quantifying their corresponding anomaly intensities.

\section{Method}
\begin{figure*}[!t]
\centering
\includegraphics[width=\textwidth]{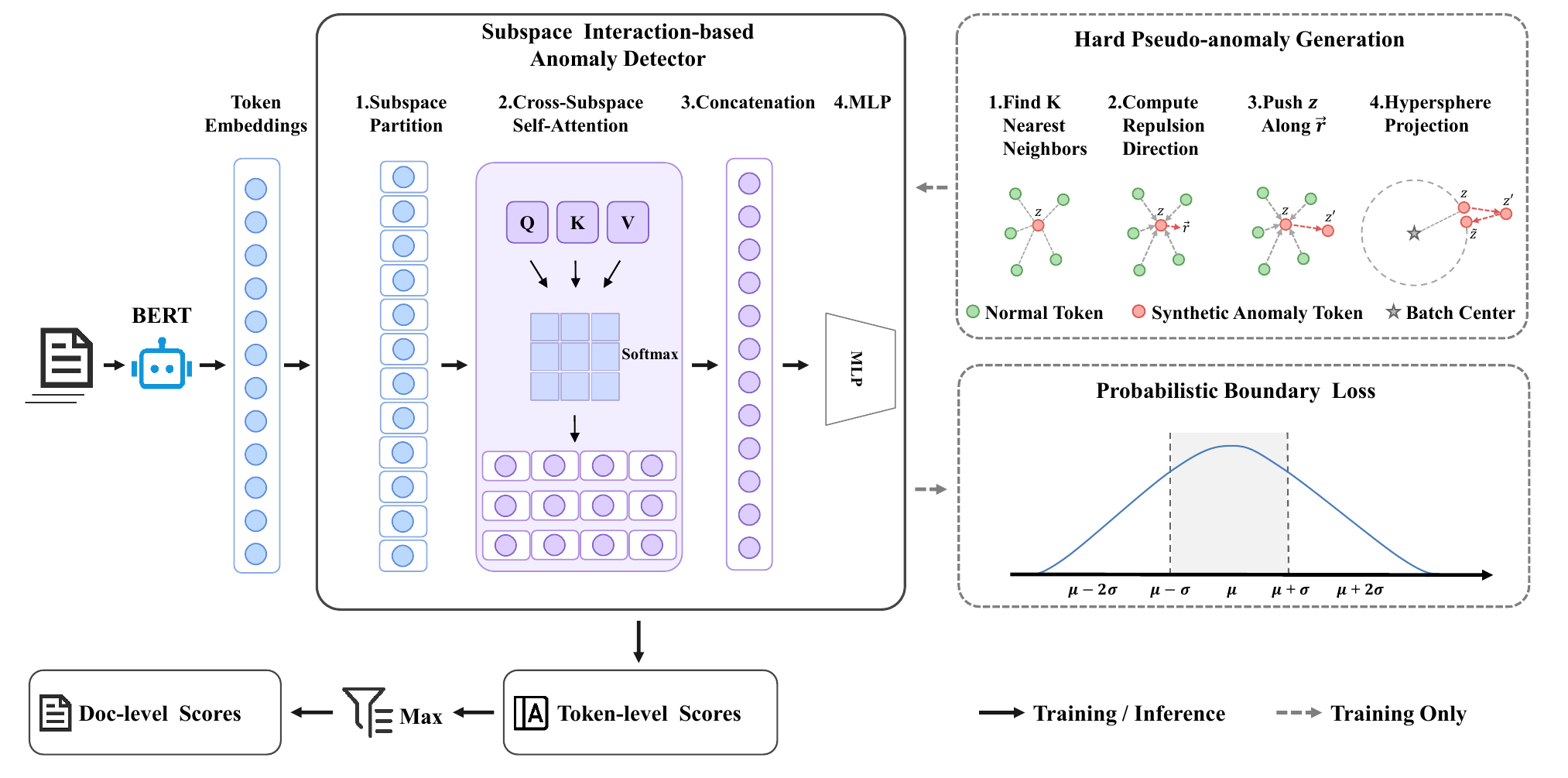}
\caption{The overall pipeline of \ourmethod for token-level text anomaly detection.}
\label{fig:pipeline}
\end{figure*}

In this section, we introduce the proposed token-level text anomaly detection framework, \ourmethod, in detail. 
As illustrated in Figure~\ref{fig:pipeline}, to capture localized anomaly signals across specific dimensions, we shift our perspective from global token embeddings to token subspaces and design a \textit{subspace interaction-based anomaly detector} (Section~\ref{method_subspace}). 
Furthermore, to mitigate the over-smoothing effect inherent in pre-trained language models (PLMs), we introduce a \textit{hard pseudo-anomaly generation} that synthesizes pseudo-anomalous tokens highly similar to normal data but lacking meaningful semantic structures (Section~\ref{method_pseudo}). 
Additionally, we introduce a \textit{probabilistic boundary loss} to ensure that the scores of anomalous tokens significantly deviate from normal instances (Section~\ref{method_loss}).

\subsection{Subspace Interaction-based Anomaly Detector}
\label{method_subspace}
Conventional text anomaly detection methods usually leverage global embeddings as the evidence to determine whether anomalies exist or not. 
However, as observed in the t-SNE visualization of token embeddings (Figure~\ref{fig:intro_tsne}), local anomalies typically do not cause significant shifts in the global representation; rather, their anomalous features are mainly concentrated in a few specific dimensions~\cite{shen2026raising}. 
Therefore, when evaluating anomaly scores directly within the global feature space, these subtle local deviations are inevitably masked and diluted by redundant normal dimensions~\cite{tan2023taming,tan2026influence}. 
To achieve effective token-level text anomaly detection, the key challenge is to address the issue of local anomaly signal dilution. 
To this end, our core idea is to shift the detection paradigm from ``coarse-grained global token representation analysis'' to ``fine-grained token subspace evaluation.'' 
Built upon this idea, we design a subspace interaction-based anomaly detector. Specifically, we partition the high-dimensional embeddings into multiple independent subspaces and model the interactive information among them. 
The ultimate goal is to enable the model to explicitly isolate anomalous subspaces and adaptively suppress irrelevant normal dimensions.

To mitigate the dilution of local anomaly signals by high-dimensional global representations, we first split the high-dimensional embeddings to capture anomaly signals at the local subspace level. In practical, we uniformly partition the input token embedding $\mathbf{z} \in \mathbb{R}^d$ into $m$ distinct subspaces, represented as:
\begin{equation}
\mathbf{z} = [\mathbf{z}^{(1)} \parallel \mathbf{z}^{(2)} \parallel \dots \parallel \mathbf{z}^{(m)}],
\end{equation}
where $\parallel$ denotes the concatenation operation, $\mathbf{z}^{(i)}$ denotes the representation of each subspace where $\mathbf{z}^{(i)} \in \mathbb{R}^{d_{sub}}$ and $d_{sub} = d/m$.

Although such partitioning enables the model to examine token representations at a finer granularity, merely partitioning the representation structurally is still insufficient to capture the underlying correlation structures among different subspaces. 
Therefore, we introduce a linear projection step inspired by the self-attention mechanism to generate Query, Key, and Value vectors for each subspace representation independently. 
This provides a unified representation space foundation for subsequent cross-subspace relationship modeling:
\begin{equation}
\mathbf{q}^{(i)} = \mathbf{z}^{(i)} \mathbf{W}_q, \quad \mathbf{k}^{(i)} = \mathbf{z}^{(i)} \mathbf{W}_k, \quad \mathbf{v}^{(i)} = \mathbf{z}^{(i)} \mathbf{W}_v,
\end{equation}
where $\mathbf{W}_q, \mathbf{W}_k, \mathbf{W}_v \in \mathbb{R}^{d_{sub} \times d_{sub}}$ are learnable projection matrices. Next, we further utilize the scaled dot-product attention mechanism to allow each subspace to aggregate information from all other subspaces, thereby explicitly modeling the relationships among different subspaces:
\begin{equation}
\mathbf{z}^{(i)}_{attn} = \sum_{j=1}^{m} \text{Softmax} \left( \frac{\mathbf{q}^{(i)} {\mathbf{k}^{(j)}}^\top}{\sqrt{d_{sub}}} \right) \mathbf{v}^{(j)}.
\end{equation}

Driven by the cross-subspace self-attention mechanism, the model can adaptively emphasize anomaly-related subspace information while suppressing redundant or irrelevant subspace information. 
More importantly, for an anomalous token, its abnormality stems not only from local feature deviations within a single subspace but also from structural inconsistencies across subspaces. 
Therefore, we further concatenate the updated $m$ subspaces back to the original dimension and feed them into a two-layer MLP to quantify this degree of inconsistency, which serves as the final token-level anomaly score:
\begin{equation}
s = \mathbf{W}_2 \text{LeakyReLU}(\mathbf{W}_1 \mathbf{z}_{attn} + b_1) + b_2,
\end{equation}
where $\mathbf{z}_{attn} = [\mathbf{z}^{(1)}_{attn} \parallel \dots \parallel \mathbf{z}^{(m)}_{attn}] \in \mathbb{R}^d$ is the representation after concatenating all updated subspace embeddings, $\mathbf{W}_1 \in \mathbb{R}^{d \times d}$ and $\mathbf{W}_2 \in \mathbb{R}^{d \times 1}$ are the weight matrices of the MLP layers, and $b_1, b_2$ are the corresponding bias terms. 
This scoring mechanism effectively prevents the local anomaly signal being diluted by operations in global redundant dimensions, allowing the final scalar score $s$ to be more directly determined by anomaly-related subspace information.

\subsection{Hard Pseudo-Anomaly Generation}
\label{method_pseudo}
Although the subspace interaction-based anomaly detector can capture anomaly signals at the subspace level, the model struggles to be trained effectively due to the lack of genuine anomaly labels under the one-class setting~\cite{xu2024calibrated}. 
Beyond this, a more intractable problem is that the optimization objective of PLMs prioritizes semantic similarity over anomaly sensitivity; hence, surface-level anomalies that preserve similar contextual meanings can still be mapped close to normal tokens in the representation space. 
This inevitably causes an over-smoothing effect on surface-level anomalies. 
To address these two issues, an intuitive approach is to artificially construct pseudo-anomalous tokens to simulate these anomalies hidden by semantic smoothing. By generating pseudo-anomalies through distance perturbation in the feature space, we construct pseudo-anomaly samples that are extremely close to the normal distribution in the feature space, yet inherently lack any actual semantic and syntactic structure. Through introducing challenging pseudo-anomaly samples into the training process, we can effectively mitigate the negative impact of the over-smoothing phenomenon, discouraging the model from relying solely on semantic similarity while encouraging it to capture subtle anomaly signals beyond global representation distance. 
Therefore, the model can better identify surface-level anomalies that are semantically close to normal tokens but violate normal semantic or syntactic structures.

To practically construct these pseudo-anomalies, we propose a distance perturbation mechanism based on local density. 
Here, the local density of each normal token is estimated by the average distance to its $K$ nearest neighbors, where a smaller average distance indicates a denser local region and thus a higher local density, and vice versa.
Considering the difference in local density among different normal tokens, we utilize the $K$-nearest neighbors algorithm to dynamically determine the perturbation direction and perturbation magnitude for each sample. 
Specifically, given a mini-batch of normal token embeddings $\{\mathbf{z}_1, \mathbf{z}_2, \dots, \mathbf{z}_B\}$, we first calculate the batch center $\boldsymbol{\mu}_B = \frac{1}{B} \sum_{i=1}^B \mathbf{z}_i$. 
Subsequently, we randomly select a proportion of normal samples to be transformed into anomalies. The proportion is denoted as $\alpha$, a hyperparameter that adjusts the pseudo-anomaly generation ratio. 
For each selected target sample $\mathbf{z}_i$, we retrieve its local $K$ nearest neighbors $\mathcal{N}_K(\mathbf{z}_i)$ within the current batch to estimate its local density and determine its perturbation direction and magnitude. 
To disrupt the original semantic anchor of the target token in the feature space without generating meaningless outliers, we aggregate all vectors pointing from the neighboring samples to $\mathbf{z}_i$ to obtain the repulsion vector $\mathbf{r}_i$ used for perturbation:
\begin{equation}
\mathbf{r}_i = \sum_{\mathbf{z}_j \in \mathcal{N}_K(\mathbf{z}_i)} (\mathbf{z}_i - \mathbf{z}_j).
\end{equation}

Next, we normalize the repulsion vector $\mathbf{r}_i$ to obtain the unit perturbation direction $\frac{\mathbf{r}_i}{\|\mathbf{r}_i\|}$. 
To ensure that the generated anomalous samples simulate subtle deviations rather than trivial out-of-distribution noise, we dynamically set the perturbation magnitude based on the local density. Specifically, we calculate the average distance between $\mathbf{z}_i$ and its $K$ neighbors as the base perturbation magnitude. Then, we apply this magnitude along the perturbation direction to $\mathbf{z}_i$, obtaining the intermediate feature $\mathbf{z}_i^{\prime}$:
\begin{equation}
\mathbf{z}_i^{\prime} = \mathbf{z}_i + \beta \cdot \frac{\mathbf{r}_i}{\|\mathbf{r}_i\|} \cdot \left( \frac{1}{K} \sum_{\mathbf{z}_j \in \mathcal{N}_K(\mathbf{z}_i)} \|\mathbf{z}_i - \mathbf{z}_j\| \right),
\end{equation}
where $\beta$ is a hyperparameter that controls the overall repulsion strength. A larger $\beta$ produces stronger perturbations, pushing the generated pseudo-anomalies farther away from their original semantic anchors.

Although the intermediate feature $\mathbf{z}_i^{\prime}$ is no longer constrained by the correct local semantics, it still risks drifting away from the global manifold.
To guarantee that the generated pseudo-anomalies consistently maintain extremely high similarity to the normal distribution in the feature space, we further apply a hypersphere projection to $\mathbf{z}_i^{\prime}$ to constrain it onto the global manifold. 
Specifically, we project $\mathbf{z}_i^{\prime}$ onto a hypersphere centered at the batch center $\boldsymbol{\mu}_B$ with a radius equal to the original distance $\|\mathbf{z}_i - \boldsymbol{\mu}_B\|$, so that the generated sample preserves the original radial distance. The final generated pseudo-anomaly sample $\tilde{\mathbf{z}}_i$ is formalized as:
\begin{equation}
\tilde{\mathbf{z}}_i = \boldsymbol{\mu}_B + (\mathbf{z}_i^{\prime} - \boldsymbol{\mu}_B) \frac{\|\mathbf{z}_i - \boldsymbol{\mu}_B\|}{\|\mathbf{z}_i^{\prime} - \boldsymbol{\mu}_B\|}.
\end{equation}
In summary, since these generated samples are entirely derived from pure physical displacement in the feature space, they inherently lack any actual semantic or syntactic structure, making them reasonable pseudo-anomalies.
Meanwhile, under the strict constraints of the hypersphere projection, they will not turn into easily identifiable extreme outliers. The discriminative signals provided by these high-quality synthesized negative samples can effectively overcome the over-smoothing effect in pre-trained representations.

\subsection{Probabilistic Boundary Loss}
\label{method_loss}
After generating pseudo-samples for training and deriving anomaly scores via the subspace interaction-based anomaly detector, the remaining question is how to design a suitable optimization objective to guide model learning. A conventional approach is to utilize the Binary Cross-Entropy (BCE) loss for supervised training. 
However, since BCE inherently relies on the empirical fitting of positive and negative samples, it is prone to overfitting to specific pseudo-anomaly generation patterns, thereby weakening generalization to unseen anomalies. 
To address this issue, we aim to guide the detector with a distribution-aware objective rather than directly fitting the generated pseudo-anomalies as a fixed positive class. 
Specifically, we map the anomaly scores into a standardized space defined by the score distribution of normal samples, so that the optimization focuses on whether a sample statistically deviates from the normal distribution instead of matching specific pseudo-anomaly patterns. 
To characterize the statistics of the standardized space, we compute the mean and variance of the anomaly scores from normal instances and standardize the scores into a Z-score format, transforming the anomaly rating into a standardized distance that deviates from the mean of the normal distribution. 
To further exploit this statistical characterization, we design a probabilistic boundary optimization objective to constrain normal samples near the distribution center while enforcing pseudo-anomalous samples to lie several standard deviations away from the distribution center.

\noindent \textbf{Score Standardization.} A primary challenge in standardizing anomaly scores is how to sample the mean and variance of normal instances. Currently, there are two main approaches, i.e., sampling from real data and sampling from a prior distribution. We ultimately opt for the prior distribution because sampling from real data exhibits significant volatility, bringing about additional training instability, and related studies show that the Gaussian distribution can well fit anomaly scores across a range of datasets, making it suitable for modeling the normal score distribution~\cite{kriegel2011interpreting,pang2019deep}. In this paper, we sample $5000$ instances from a standard normal distribution to obtain the reference mean $\mu_{ref}$ and standard deviation $\sigma_{ref}$ for the anomaly scores of normal samples. Then, we convert the raw anomaly score $s$ of each token into a standard Z-score:
\begin{equation}
dev(s) = \frac{s - \mu_{ref}}{\sigma_{ref}},
\end{equation}
which maps the raw anomaly scores into a standardized statistical space. Here, $dev(s)$ explicitly quantifies the number of standard deviations by which a token's score deviates from the normal data center, thereby providing a statistically interpretable measure of anomaly degree.

\noindent \textbf{Probabilistic Boundary Optimization.} 
To strengthen the generalization ability of the model, we design a probabilistic boundary optimization objective for training. 
It encourages the detector to learn anomaly scores based on statistically meaningful deviations from the normal distribution, rather than overfitting to specific pseudo-anomaly patterns.
For a mini-batch of size $B$, the overall probabilistic boundary loss $\mathcal{L}_{PBL}$ is calculated as follows:
\begin{equation}
\mathcal{L}_{PBL} = \frac{1}{B} \sum_{i=1}^{B} \left[ (1 - y_i) |dev(s_i)| + y_i \max(0, a - dev(s_i)) \right],
\end{equation}
where $y_i = 1$ denotes a pseudo-anomalous sample, $y_i = 0$ denotes a normal sample, and $a$ is the confidence boundary parameter for the Z-score. This loss function forces the scores of normal tokens to tightly cluster around the distribution center (i.e., $|dev(s)| \to 0$), while enforcing the generated anomalous samples to stay at least $a$ standard deviations away from the normal center, thereby establishing a clear probabilistic boundary between normal and pseudo-anomalous samples.

In summary, the probabilistic boundary loss transforms the originally uninterpretable anomaly scores into a standardized distance metric with clear statistical significance and formulates an optimization objective constrained by a confidence boundary. This design not only avoids overfitting to the pseudo-anomaly distribution but also better guides the model in learning robust anomaly scores.

\noindent \textbf{Document-Level Score Aggregation.} 
After obtaining token-level anomaly scores, it is necessary to aggregate them into a document-level score for document-level anomaly detection~\cite{cao2026towards}. 
Specifically, given a document $\mathcal{X}_i$ containing $T_i$ tokens, the token-level anomaly detector assigns an anomaly score $s_{i,t}$ to each token $t$. 
The key question is how to aggregate these token-level scores into an overall document-level anomaly score.

Existing methods typically employ mean pooling for aggregation since raw scores are inherently unstable and vulnerable to noise, and averaging can yield a more robust overall estimation~\cite{cao2026towards}. In practice, however, we argue that this approach is highly suboptimal for fine-grained detection tasks, and we instead utilize max pooling as an alternative aggregation function to address the limitations of mean pooling. The rationale is that in token-level anomaly detection datasets, the proportion of anomalous tokens is extremely small, often accounting for less than $1\%$. Under these circumstances, mean pooling inevitably averages the few strong local anomaly signals with the overwhelming majority of normal tokens, causing the critical anomaly information to be significantly diluted. Therefore, we adopt max pooling as the aggregation function, and the document-level anomaly score $S_i$ for document $\mathcal{X}_i$ is calculated as follows:
\begin{equation}
    S_i = \max_{1 \le t \le T_i} s_{i,t}.
\end{equation}
This design allows the document-level score to be dominated by the most suspicious token-level evidence, thereby preserving sparse but critical anomaly signals that would otherwise be smoothed out by averaging. 

\begin{table*}[t]
\centering
\caption{Main results on AUROC and AUPRC for three datasets. Best results are highlighted in bold and shaded.}
\label{tab:main_results}
\renewcommand{\arraystretch}{1}
\begin{tabular}{
>{\centering\arraybackslash}p{2.0cm}| 
>{\raggedright\arraybackslash}p{2.2cm}|
>{\centering\arraybackslash}p{1.2cm} >{\centering\arraybackslash}p{1.2cm}|
>{\centering\arraybackslash}p{1.2cm} >{\centering\arraybackslash}p{1.2cm}|
>{\centering\arraybackslash}p{1.2cm} >{\centering\arraybackslash}p{1.2cm}|
>{\centering\arraybackslash}p{1.2cm} >{\centering\arraybackslash}p{1.2cm}}
\toprule
\multirow{2}{*}{\textbf{Level}} 
& \multirow{2}{*}{\textbf{Methods}}
& \multicolumn{2}{c|}{\textbf{SMS\_Spam}}
& \multicolumn{2}{c|}{\textbf{Review}}
& \multicolumn{2}{c|}{\textbf{Grammar}}
& \multicolumn{2}{c}{\textbf{Average}} \\
\cline{3-10}
& & \rule{0pt}{1.2em} AUROC & AUPRC
& AUROC & AUPRC
& AUROC & AUPRC
& AUROC & AUPRC \\
\midrule

& LOF
& 44.31 & 0.77
& 66.35 & 1.99
& 52.26 & 2.88
& 54.31 & 1.88 \\
& iForest
& 76.18 & 2.10
& 67.57 & 6.70
& 41.91 & 2.37
& 61.89 & 3.72 \\
& ECOD
& 82.57 & 2.86
& 69.49 & 7.32
& 47.15 & 2.61
& 66.40 & 4.26 \\
& DeepSVDD
& 58.74 & 1.10
& 59.88 & 2.20
& 50.52 & 3.00
& 56.38 & 2.10 \\
& AutoEncoder
& 36.64 & 0.67
& 63.01 & 1.74
& 63.01 & 3.68
& 54.22 & 2.03 \\
& LUNAR
& 66.22 & 1.26
& 81.78 & 4.49
& 60.80 & 3.47
& 69.60 & 3.07 \\
& TokenCore
& 70.26 & 1.43
& \cellcolor{gray!20}\textbf{81.89} & 4.81
& 64.00 & 3.80
& 72.05 & 3.35 \\
& GPT-4.1-nano
& 92.19 & \cellcolor{gray!20}\textbf{38.79}
& 47.48 & 7.36
& 42.41 & 2.75
& 60.69 & 16.30 \\
\multirow{-9}{1.4cm}{\centering\textbf{Token-level}}
& \ourmethod
& \cellcolor{gray!20}\textbf{98.44} & 28.74
& 74.78 & \cellcolor{gray!20}\textbf{17.82}
& \cellcolor{gray!20}\textbf{72.14} & \cellcolor{gray!20}\textbf{5.22}
& \cellcolor{gray!20}\textbf{81.79} & \cellcolor{gray!20}\textbf{17.26} \\
\midrule

& LOF
& 55.72 & 18.49
& 91.42 & 43.83
& 57.55 & 19.91
& 68.23 & 27.41 \\
& iForest
& 45.98 & 14.40
& 87.55 & 33.98
& 60.85 & 22.82
& 64.79 & 23.73 \\
& ECOD
& 47.10 & 14.75
& 87.07 & 31.34
& 61.88 & 23.90
& 65.35 & 23.33 \\
& DeepSVDD
& 41.12 & 13.17
& 78.93 & 20.95
& 65.32 & 28.68
& 61.79 & 20.93 \\
& AutoEncoder
& 46.43 & 14.56
& 88.45 & 37.36
& 68.07 & 27.72
& 67.65 & 26.55 \\
& LUNAR
& 58.78 & 19.14
& 95.66 & 65.04
& 67.05 & 26.72
& 73.83 & 36.97 \\
& TokenCore
& 60.59 & 20.06
& 95.68 & 65.10
& 64.51 & 25.69
& 73.59 & 36.95 \\
& GPT-4.1-nano
& 54.99 & 16.48
& 43.04 & 9.12
& 62.05 & 23.82
& 53.36 & 16.47 \\
\multirow{-9}{1.4cm}{\centering\textbf{Doc-level}}
& \ourmethod
& \cellcolor{gray!20}\textbf{88.53} & \cellcolor{gray!20}\textbf{48.28}
& \cellcolor{gray!20}\textbf{95.75} & \cellcolor{gray!20}\textbf{83.13}
& \cellcolor{gray!20}\textbf{70.10} & \cellcolor{gray!20}\textbf{35.35}
& \cellcolor{gray!20}\textbf{84.79} & \cellcolor{gray!20}\textbf{55.59} \\
\bottomrule
\end{tabular}
\end{table*}

\subsection{Computational Complexity}
We analyze the computational complexity of \ourmethod for a document with $T$ tokens and embedding dimension $d$, assuming a training mini-batch size of $B$ tokens. Consistent with standard protocols, token-level embeddings are extracted once via the pretrained language model and cached~\cite{cao2026towards}, excluding the backbone forward pass from our analysis. The subspace interaction-based anomaly detector first partitions embeddings into $m$ subspaces and performs cross-subspace attention, costing $O(Td^2/m + Tmd)$, and subsequently employs a two-layer MLP for token-level scoring with complexity $O(Td^2)$. The hard pseudo-anomaly generation is exclusively activated during training; it computes the batch center and conducts in-batch $K$-nearest neighbor retrieval and neighborhood repulsion, incurring a training-only overhead of $O(B^2d + BKd)$ per mini-batch. Finally, both score standardization and document-level max-pooling aggregation scale linearly, costing $O(T)$ per document. Consequently, the overall inference complexity per document scales as $O(Td^2 + Tmd)$. Because $m \ll d$ in practice, the overhead is dominated by the linear projections and MLP layers, which remain strictly linear with respect to the document length $T$, ensuring that \ourmethod is highly efficient for practical deployment.

\section{Experiments}
\subsection{Experiment Setup}
\noindent\textbf{Datasets.} We conduct experiments on three public benchmark datasets encompassing diverse anomaly patterns, including a grammatical error dataset (Grammar), a negative sentiment dataset (Review), and a text corruption dataset (SMS\_Spam). Following the standard protocol of TokenCore~\cite{cao2026towards}, we allocate 50\% of the normal instances for training. The remaining 50\% of the normal instances, together with all anomalous instances, form the test set.

\noindent\textbf{Baselines.} We compare \ourmethod with representative baselines, including LOF~\cite{breunig2000lof}, iForest~\cite{liu2008isolation}, ECOD~\cite{li2022ecod}, DeepSVDD~\cite{ruff2018deep}, AutoEncoder~\cite{zhou2017anomaly}, LUNAR~\cite{goodge2022lunar}, and TokenCore~\cite{cao2026towards}. To evaluate the performance of Large Language Model (LLM) on token-level text anomaly detection, we include GPT-4.1-nano\footnote{\url{https://developers.openai.com/api/docs/models/gpt-4.1-nano}} as an additional baseline.

\noindent\textbf{Evaluation and Implementation.} We report AUROC and AUPRC as the main metrics. For all methods, we report the average results across 3 random seeds. All methods use embeddings extracted from BERT-base-uncased\footnote{\url{https://huggingface.co/google-bert/bert-base-uncased}}~\cite{devlin2019bert}. Since the tokenizer typically splits a word into multiple subwords (e.g., splitting ``playing'' into ``play'' and ``\#\#ing''), we apply max pooling to aggregate the subword embeddings, obtaining word-level representations that align with token annotations~\cite{cao2026towards}.

\subsection{Main Results}
Table~\ref{tab:main_results} reports the comparison results in terms of AUROC and AUPRC. We have the following observations.
\ding{182}~At the document level, \ourmethod achieves the best AUROC and AUPRC across all three datasets. Compared to the strongest baseline, our method obtains a relative performance gain of over 10\%. This demonstrates its consistent and effective performance for text anomaly detection under diverse anomaly patterns.
\ding{183}~At the token level, \ourmethod obtains the best average AUROC and AUPRC. Particularly on the SMS\_Spam dataset, it reaches an impressive AUROC of 98.44, even outperforming GPT-4.1-nano (92.19). This demonstrates that our model is capable of capturing subtle anomaly signals, thereby proving its effectiveness in token-level text anomaly detection.
\ding{184}~Existing methods (such as TokenCore and LUNAR) demonstrate competitive performance on certain specific datasets; however, their performance varies drastically across different datasets. This reflects the inherent characteristics of pre-trained language models (PLMs) like BERT, which are primarily optimized for semantic similarity. In the Review dataset, semantic anomalies are distinctly separated within the embedding space, enabling various methods to achieve high detection performance. However, for text corruptions in SMS\_Spam and grammatical errors in Grammar, embeddings from PLMs inevitably smooth out these surface-level anomalies, leading to suboptimal detection results.
\ding{185}~Notably, the LLM baseline generally underperforms at both token and document levels. Although GPT-4.1-nano demonstrates certain zero-shot capabilities on the SMS\_Spam dataset, its overall performance still lags significantly behind. This highlights the necessity of designing specialized architectures tailored for token-level text anomaly detection.

\begin{table}[t]
\centering
\caption{Ablation results on AUROC for three datasets. Best results are highlighted in bold and shaded.}
\label{tab:ablation_results}
\renewcommand{\arraystretch}{1}
\begin{tabular}{
>{\centering\arraybackslash}p{0.7cm}| 
>{\raggedright\arraybackslash}p{1.8cm}|
>{\centering\arraybackslash}p{1.4cm}|
>{\centering\arraybackslash}p{1.4cm}|
>{\centering\arraybackslash}p{1.4cm}}
\toprule
\textbf{Level} & \textbf{Variants} & \textbf{SMS\_Spam} & \textbf{Review} & \textbf{Grammar} \\
\midrule

\multirow{4}{*}{\rotatebox{90}{\textbf{Token}}}
& \ourmethod
& \cellcolor{gray!20}\textbf{98.44} & \cellcolor{gray!20}\textbf{74.78} & \cellcolor{gray!20}\textbf{72.14} \\
& w/o Sub-Int
& 49.39 & 71.35 & 59.77 \\
& w/o Hard-Gen
& 49.37 & 43.19 & 48.66 \\
& w/o Prob-Loss
& 60.60 & 59.50 & 62.86 \\
\midrule

\multirow{4}{*}{\rotatebox{90}{\textbf{Doc}}}
& \ourmethod
& \cellcolor{gray!20}\textbf{88.53} & \cellcolor{gray!20}\textbf{95.75} & \cellcolor{gray!20}\textbf{70.10} \\
& w/o Sub-Int
& 45.62 & 95.23 & 57.56 \\
& w/o Hard-Gen
& 62.89 & 73.16 & 62.46 \\
& w/o Prob-Loss
& 54.75 & 92.68 & 57.58 \\
\bottomrule
\end{tabular}
\end{table}
\subsection{Ablation Study} 
To validate the key components of \ourmethod, we evaluate three variants: \ding{182} \textbf{w/o Sub-Int}, which replaces the cross-subspace self-attention mechanism with an MLP; \ding{183} \textbf{w/o Hard-Gen}, which uses Gaussian noise instead of distance perturbation to generate pseudo-anomalies; and \ding{184} \textbf{w/o Prob-Loss}, which trains the anomaly detector with Binary Cross-Entropy (BCE) loss instead of probabilistic boundary loss.

The results are summarized in Table~\ref{tab:ablation_results}, which shows that all components consistently contribute to the final performance. \ding{182} \textbf{w/o Sub-Int} causes a consistent performance drop. This indicates that treating high-dimensional token embeddings as indivisible black boxes severely dilutes local anomaly signals with redundant normal feature dimensions. \ding{183} \textbf{w/o Hard-Gen} yields the largest drop across most settings, e.g., $72.14 \rightarrow 48.66$ on Grammar at the token level. This shows that directionless Gaussian noise fails to generate effective hard pseudo-anomalies to counteract the over-smoothing effect inherent in PLMs, making it difficult for the model to break free from its reliance on semantic similarity. \ding{184} \textbf{w/o Prob-Loss} results in a severe decrease, especially when identifying text gibberish, e.g., $88.53 \rightarrow 54.75$ on SMS\_Spam at the document level. Without a statistical boundary constructed from the mean and variance of normal data, the standard BCE objective easily overfits to pseudo-anomalies, limiting its generalization to unseen anomalies.

\begin{figure}[t]
\centering
\subfloat[Grammar]{
\includegraphics[width=0.31\columnwidth]{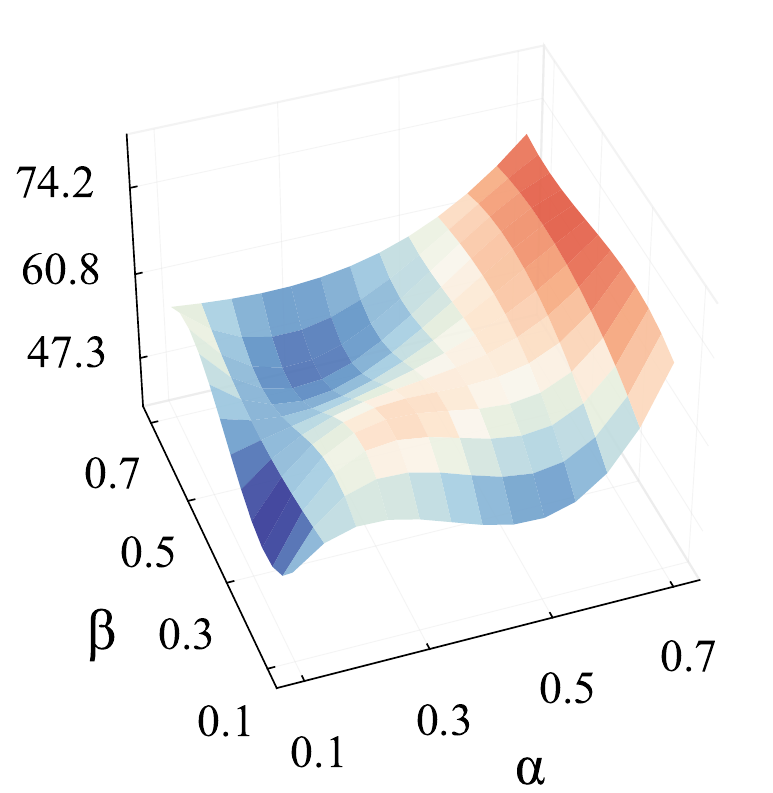}}
\hfill
\subfloat[Review]{
\includegraphics[width=0.31\columnwidth]{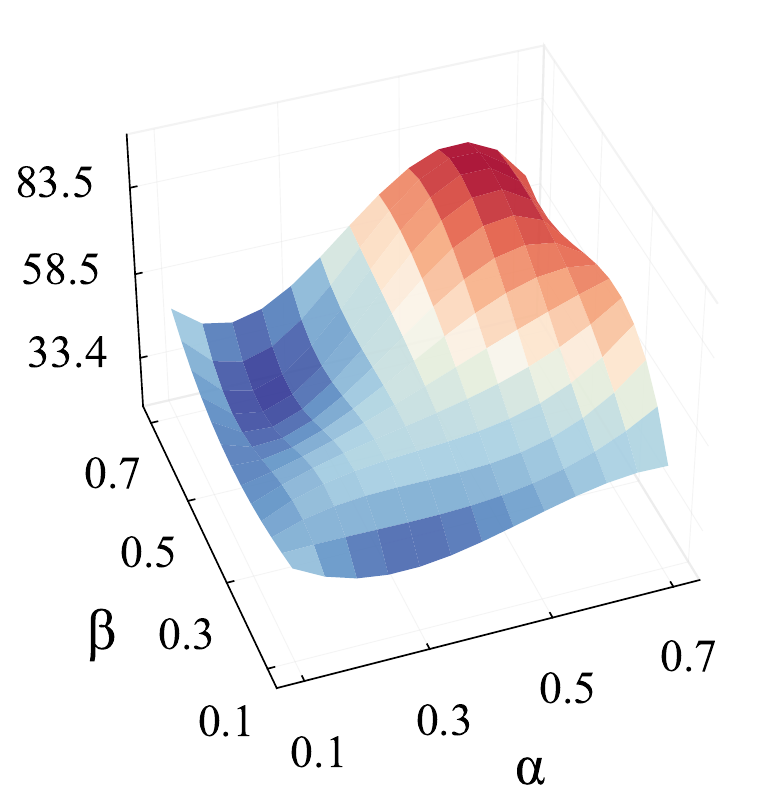}}
\hfill
\subfloat[SMS\_Spam]{
\includegraphics[width=0.31\columnwidth]{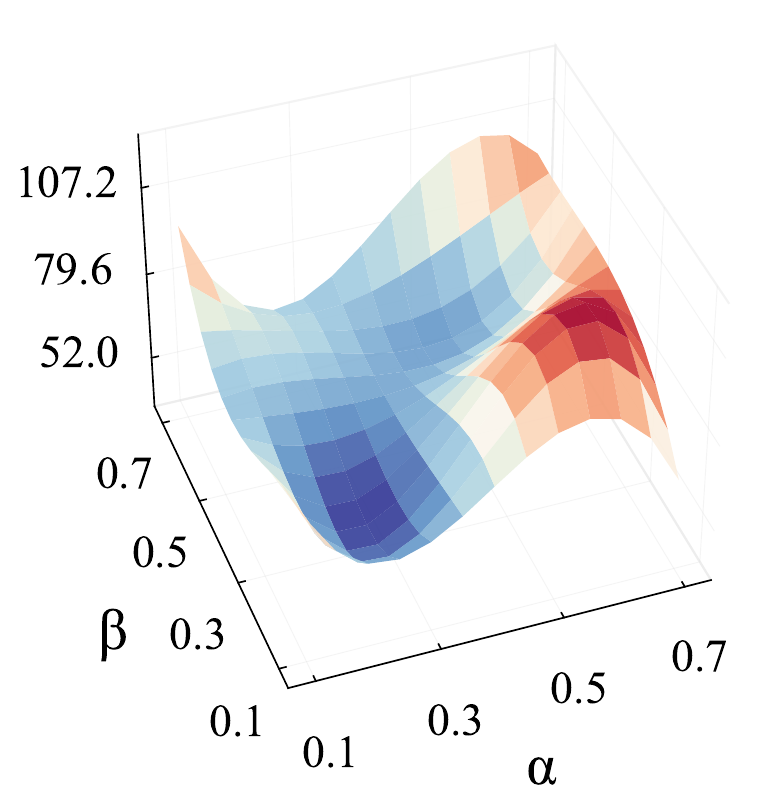}}
\caption{Hyperparameter sensitivity analysis of token-level AUROC.}
\label{fig:hyper_token}
\end{figure}

\begin{figure}[t]
\centering
\subfloat[Grammar]{
\includegraphics[width=0.31\columnwidth]{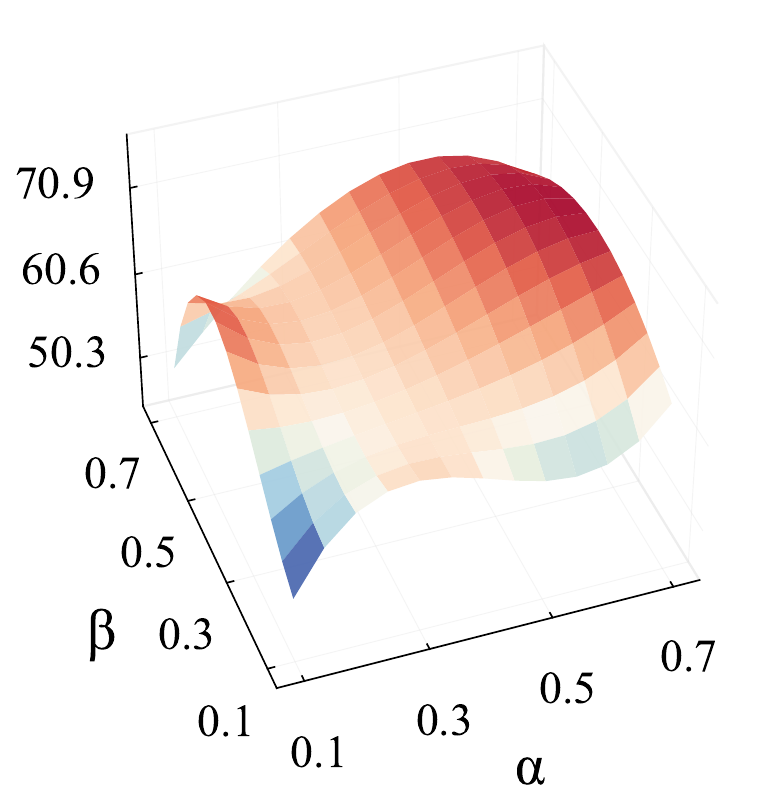}}
\hfill
\subfloat[Review]{
\includegraphics[width=0.31\columnwidth]{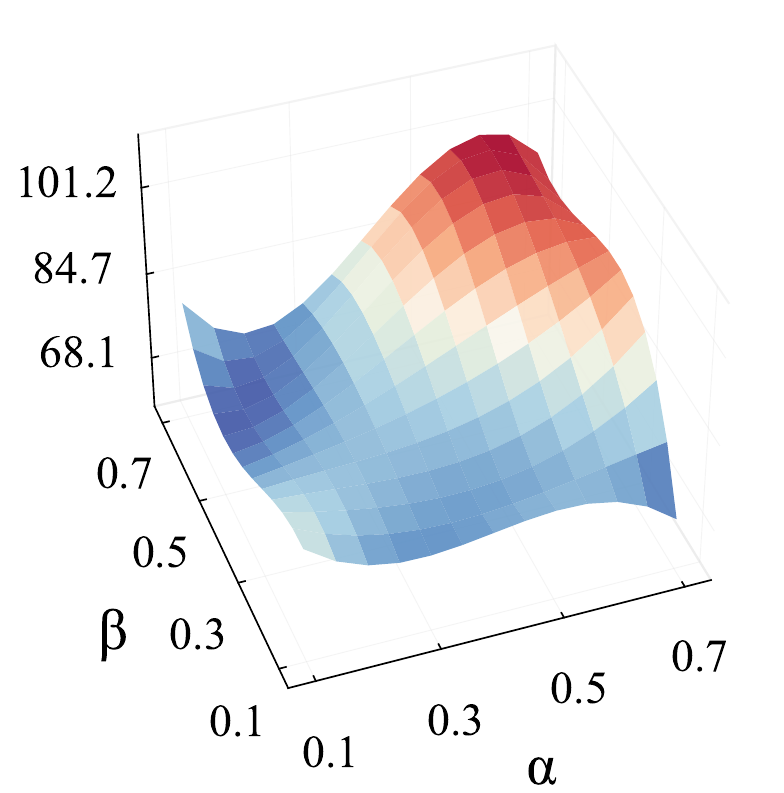}}
\hfill
\subfloat[SMS\_Spam]{
\includegraphics[width=0.31\columnwidth]{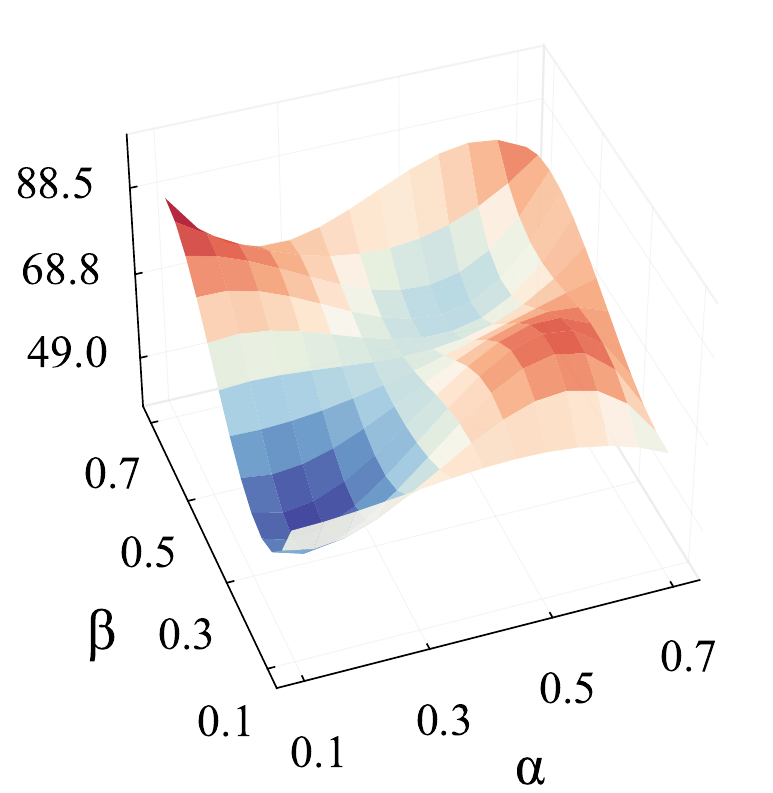}}
\caption{Hyperparameter sensitivity analysis of document-level AUROC.}
\label{fig:hyper_doc}
\end{figure}
\subsection{Hyperparameter Analysis}
We study the sensitivity of \ourmethod to the hard pseudo-anomaly generation parameters, the anomaly ratio $\alpha$ and perturbation scale $\beta$, as shown in Figures~\ref{fig:hyper_token} and~\ref{fig:hyper_doc}. Across all datasets, optimal performance is consistently achieved with a relatively large $\alpha$, demonstrating the benefit of pseudo-anomaly samples for model training. Furthermore, the optimal perturbation scale $\beta$ depends on the specific anomaly type. On Grammar and Review, a larger $\beta$ is preferred, as a small $\beta$ produces negligible noise that fails to disrupt the original semantic and syntactic structures. Conversely, SMS\_Spam favors a smaller $\beta$, as an excessively large $\beta$ pushes generated tokens into overly distant feature-space regions, turning them into trivial outliers rather than \textit{hard} pseudo-anomalies and degrading detection precision.

\subsection{Efficiency Analysis}
\begin{figure}[t]
\centering
\subfloat[]{
\includegraphics[width=0.5\columnwidth]{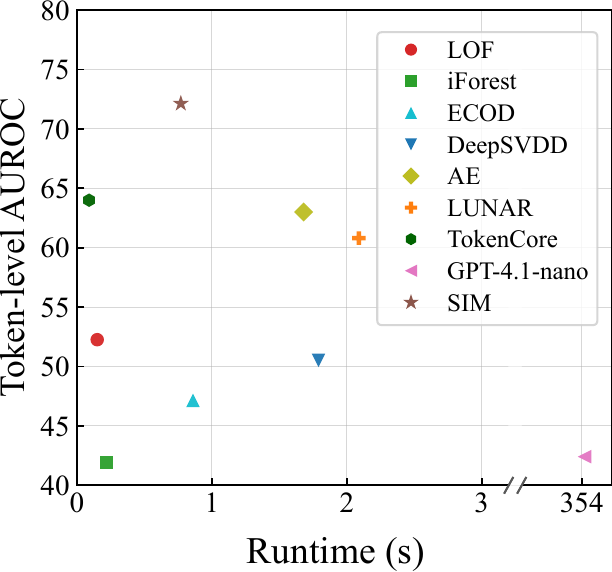}
\label{fig:efficiency}
}
\subfloat[]{
\includegraphics[width=0.5\columnwidth]{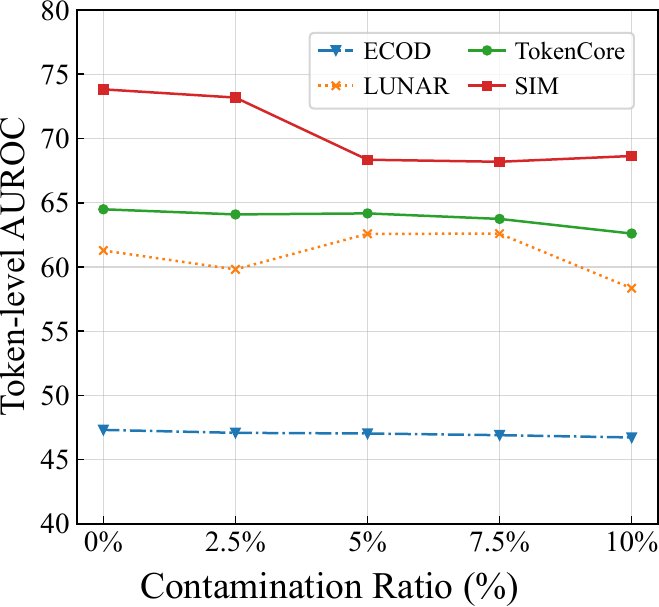}
\label{fig:robustness}
}
\caption{Performance evaluation on the Grammar dataset. (a) Efficiency analysis comparing various methods in terms of token-level AUROC and runtime. (b) Robustness analysis assessing AUROC stability under varying training data contamination ratios.}
\end{figure}

To assess the runtime efficiency and performance trade-off of \ourmethod, Figure~\ref{fig:efficiency} compares token-level AUROC and total runtime with representative baselines on Grammar. Overall, \ourmethod achieves the highest AUROC with consistently low runtime. Specifically, \ourmethod reduces execution time by an order of magnitude compared to GPT-4.1-nano and is substantially faster than AE, DeepSVDD, and LUNAR, achieving more than a twofold speedup with higher detection accuracy. Notably, despite being a deep learning-based model, \ourmethod (0.77s) is faster than the training-free traditional baseline ECOD (0.86s). Although TokenCore and LOF are slightly faster, their detection performance is substantially inferior; for example, \ourmethod outperforms TokenCore by 8.14\% in token-level AUROC with negligible efficiency trade-off. These results demonstrate that \ourmethod achieves a favorable balance between accuracy and efficiency without prohibitive computational overhead.

\subsection{Robustness Analysis}
To evaluate the robustness of \ourmethod against contaminated training data, we artificially corrupt varying proportions of normal token embeddings with zero-mean Gaussian noise scaled to three times the dataset's standard deviation, as shown in Figure~\ref{fig:robustness}. Overall, as the contamination ratio increases, the detection performance of all models exhibits a predictable downward trend. However, \ourmethod preserves strong performance under increasing contamination. Most notably, even under the most severe scenario (10\% contamination), the degraded performance of \ourmethod still significantly surpasses the peak performance of all baselines in the strictly clean (0\% contamination) setting, maintaining a clear margin over strong baselines. Ultimately, these results compellingly demonstrate that \ourmethod can effectively mitigate the adverse impact of the inevitable data contamination found in real-world scenarios.

\begin{figure}[t]
    \centering
    \includegraphics[width=0.48\textwidth]{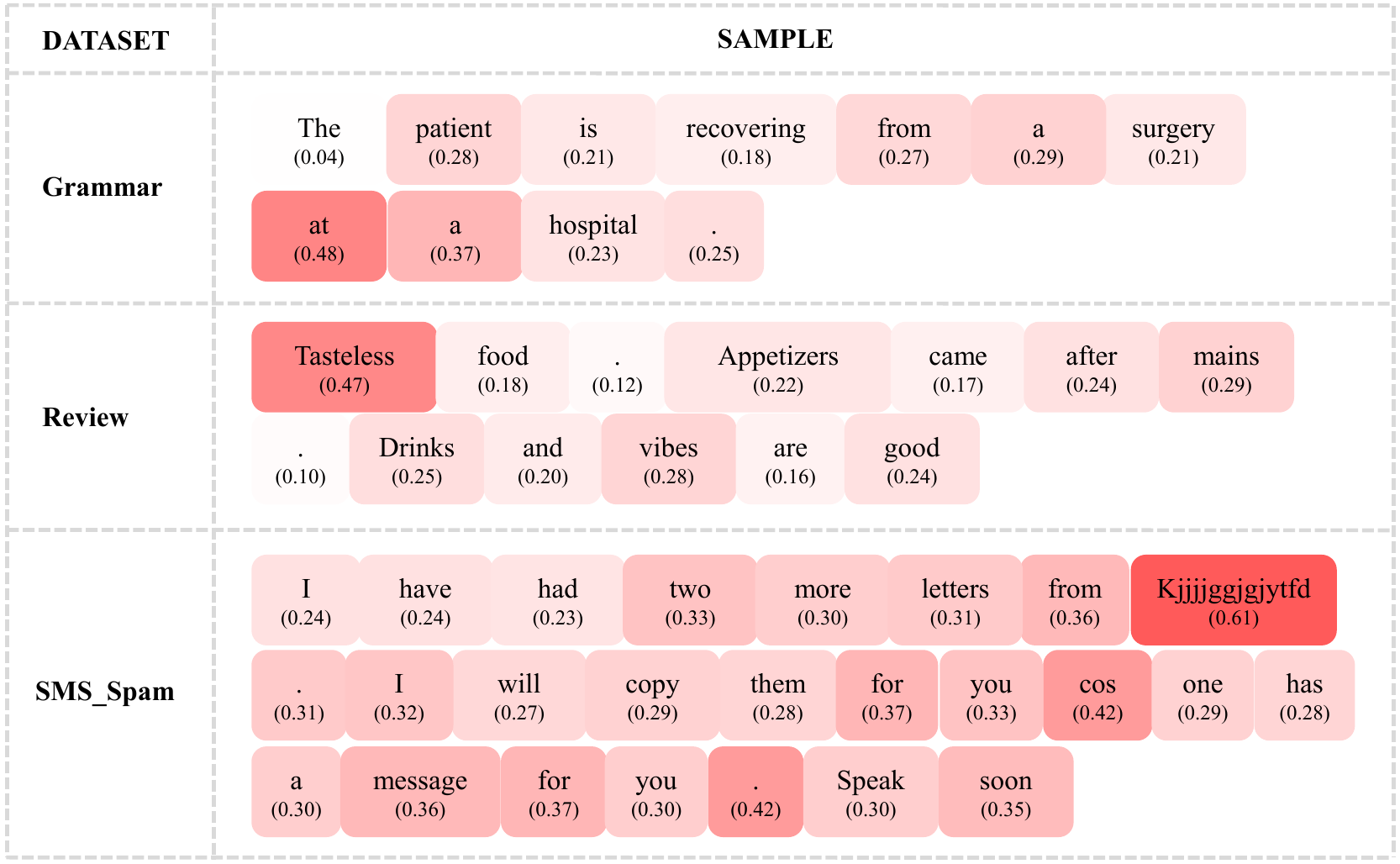}
    \caption{Interpretability visualization of token-level anomaly attributions across different datasets. Darker shading corresponds to higher anomaly scores.}
    \label{fig:casestudy}
\end{figure}
\subsection{Interpretability Analysis}
To investigate the interpretability of our proposed framework, we visualize the token-level anomaly scores generated by \ourmethod across three distinct datasets in Figure~\ref{fig:casestudy}. The visualizations reveal that \ourmethod effectively mitigates the over-smoothing effect of PLMs and successfully captures fine-grained local anomaly signals. For instance, in the Grammar dataset, the model overcomes latent representation smoothing to precisely pinpoint the surface anomaly preposition ``at'' (score: 0.48), the exact token disrupting the sentence structure. In the Review dataset, it accurately assigns a high score to the strong negative descriptor ``Tasteless'' (0.47), while successfully ignoring neutral semantic tokens such as ``food'' and ``Drinks''. Similarly, in the SMS\_Spam dataset, the model effectively detects local structural corruption by assigning the highest score (0.61) to the anomalous string ``Kjjjjggjgjytfd'', thereby accurately pinpointing this irregular character sequence.

\section{Conclusion}
In this paper, we focus on token-level text anomaly detection, aiming to identify anomalous tokens within a document and provide fine-grained anomaly localization results. To this end, we propose \ourmethod, a Subspace Interaction-based Method for token-level text anomaly detection. To address the dilution of local anomaly signals, \ourmethod adopts a subspace interaction-based anomaly detector to dynamically amplify localized anomaly signals hidden in specific dimensions. To counteract the inherent over-smoothing effect of pre-trained language models (PLMs), we introduce hard pseudo-anomaly generation to construct pseudo-anomalous tokens and design a probabilistic boundary loss to standardize anomaly scores into statistical distances, enforcing anomalous instances to deviate from the normal distribution center. Extensive experiments demonstrate that \ourmethod achieves state-of-the-art performance at both token and document levels on multiple benchmark datasets, with remarkable efficiency, robustness, and interpretability.

\section*{Acknowledgment}
The work of Qingfeng Chen was partially supported by the Specific Research Project of Guangxi for Research Bases and Talents under Grant No. GuiKe AD24010011 and the Key Research \& Development Program Project of Guangxi under Grant No. GuiKe AB25069095. The work of Kehan Yan was partially supported by the Innovation Project of Guangxi Graduate Education under Grant No. YCSW2026145.


\bibliographystyle{IEEEtran}
\bibliography{IEEEtran}

@inproceedings{cao2026towards,
  title={Towards Token-Level Text Anomaly Detection},
  author={Cao, Yang and Yu, Bicheng and Yang, Sikun and Liu, Ming and Yang, Yujiu},
  booktitle={Proceedings of the ACM Web Conference 2026},
  pages={8733--8736},
  year={2026}
}

@article{li2020survey,
  title={A survey on deep learning for named entity recognition},
  author={Li, Jing and Sun, Aixin and Han, Jianglei and Li, Chenliang},
  journal={IEEE transactions on knowledge and data engineering},
  volume={34},
  number={1},
  pages={50--70},
  year={2020},
  publisher={IEEE}
}

@inproceedings{pang2019deep,
  title={Deep anomaly detection with deviation networks},
  author={Pang, Guansong and Shen, Chunhua and Van Den Hengel, Anton},
  booktitle={Proceedings of the 25th ACM SIGKDD international conference on knowledge discovery \& data mining},
  pages={353--362},
  year={2019}
}

@article{han2022adbench,
  title={Adbench: Anomaly detection benchmark},
  author={Han, Songqiao and Hu, Xiyang and Huang, Hailiang and Jiang, Minqi and Zhao, Yue},
  journal={Advances in neural information processing systems},
  volume={35},
  pages={32142--32159},
  year={2022}
}

@inproceedings{zhao2025freegad,
  title={Freegad: A training-free yet effective approach for graph anomaly detection},
  author={Zhao, Yunfeng and Liu, Yixin and Li, Shiyuan and Chen, Qingfeng and Zheng, Yu and Pan, Shirui},
  booktitle={Proceedings of the 34th ACM International Conference on Information and Knowledge Management},
  pages={4379--4389},
  year={2025}
}

@inproceedings{pan2023prem,
  title={PREM: A Simple Yet Effective Approach for Node-Level Graph Anomaly Detection},
  author={Pan, J and Liu, Y and Zheng, Y and Pan, S},
  booktitle={2023 IEEE International Conference on Data Mining (ICDM)},
  year={2023},
  organization={IEEE}
}

@inproceedings{yin2024mcm,
  title={Mcm: Masked cell modeling for anomaly detection in tabular data},
  author={Yin, Jiaxin and Qiao, Yuanyuan and Zhou, Zitang and Wang, Xiangchao and Yang, Jie},
  booktitle={The Twelfth International Conference on Learning Representations},
  year={2024}
}

@inproceedings{ye2025drl,
  title={DRL: Decomposed representation learning for tabular anomaly detection},
  author={Ye, Hangting and Zhao, He and Fan, Wei and Zhou, Mingyuan and Guo, Dandan and Chang, Yi},
  booktitle={International Conference on Learning Representations},
  volume={2025},
  pages={24554--24589},
  year={2025}
}

@inproceedings{shi2022revisiting,
  title={Revisiting Over-smoothing in BERT from the Perspective of Graph},
  author={Shi, H and GAO, J and Xu, H and Liang, X and Li, Z and Kong, L and Lee, SMS and Kwok, J},
  booktitle={International Conference on Learning Representations},
  year={2022}
}

@article{tu2024weighted,
  title={Weighted subspace anomaly detection in high-dimensional space},
  author={Tu, Jiankai and Liu, Huan and Li, Chunguang},
  journal={Pattern Recognition},
  volume={146},
  pages={110056},
  year={2024},
  publisher={Elsevier}
}

@article{pang2021deep,
  title={Deep learning for anomaly detection: A review},
  author={Pang, Guansong and Shen, Chunhua and Cao, Longbing and Hengel, Anton Van Den},
  journal={ACM computing surveys (CSUR)},
  volume={54},
  number={2},
  pages={1--38},
  year={2021},
  publisher={ACM New York, NY, USA}
}

@article{cao2025anomaly,
  title={Anomaly detection based on isolation mechanisms: A survey},
  author={Cao, Yang and Xiang, Haolong and Zhang, Hang and Zhu, Ye and Ting, Kai Ming},
  journal={Machine Intelligence Research},
  volume={22},
  number={5},
  pages={849--865},
  year={2025},
  publisher={Springer}
}

@article{chalapathy2019deep,
  title={Deep learning for anomaly detection: A survey},
  author={Chalapathy, Raghavendra and Chawla, Sanjay},
  journal={arXiv preprint arXiv:1901.03407},
  year={2019}
}

@inproceedings{breunig2000lof,
  title={LOF: identifying density-based local outliers},
  author={Breunig, Markus M and Kriegel, Hans-Peter and Ng, Raymond T and Sander, J{\"o}rg},
  booktitle={Proceedings of the 2000 ACM SIGMOD international conference on Management of data},
  pages={93--104},
  year={2000}
}

@inproceedings{liu2008isolation,
  title={Isolation forest},
  author={Liu, Fei Tony and Ting, Kai Ming and Zhou, Zhi-Hua},
  booktitle={2008 eighth ieee international conference on data mining},
  pages={413--422},
  year={2008},
  organization={IEEE}
}

@article{li2022ecod,
  title={Ecod: Unsupervised outlier detection using empirical cumulative distribution functions},
  author={Li, Zheng and Zhao, Yue and Hu, Xiyang and Botta, Nicola and Ionescu, Cezar and Chen, George H},
  journal={IEEE Transactions on Knowledge and Data Engineering},
  volume={35},
  number={12},
  pages={12181--12193},
  year={2022},
  publisher={IEEE}
}

@inproceedings{zhou2017anomaly,
  title={Anomaly detection with robust deep autoencoders},
  author={Zhou, Chong and Paffenroth, Randy C},
  booktitle={Proceedings of the 23rd ACM SIGKDD international conference on knowledge discovery and data mining},
  pages={665--674},
  year={2017}
}

@inproceedings{ruff2018deep,
  title={Deep one-class classification},
  author={Ruff, Lukas and Vandermeulen, Robert and Goernitz, Nico and Deecke, Lucas and Siddiqui, Shoaib Ahmed and Binder, Alexander and M{\"u}ller, Emmanuel and Kloft, Marius},
  booktitle={International conference on machine learning},
  pages={4393--4402},
  year={2018},
  organization={PMLR}
}

@inproceedings{goodge2022lunar,
  title={Lunar: Unifying local outlier detection methods via graph neural networks},
  author={Goodge, Adam and Hooi, Bryan and Ng, See-Kiong and Ng, Wee Siong},
  booktitle={Proceedings of the AAAI conference on artificial intelligence},
  volume={36},
  number={6},
  pages={6737--6745},
  year={2022}
}

@article{cao2025tad,
  title={Tad-bench: A comprehensive benchmark for embedding-based text anomaly detection},
  author={Cao, Yang and Yang, Sikun and Li, Chen and Xiang, Haolong and Qi, Lianyong and Liu, Bo and Li, Rongsheng and Liu, Ming},
  journal={arXiv preprint arXiv:2501.11960},
  year={2025}
}

@article{cao2025text,
  title={Text Anomaly Detection with Simplified Isolation Kernel},
  author={Cao, Yang and Yang, Sikun and Yang, Yujiu and Qi, Lianyong and Liu, Ming},
  journal={arXiv preprint arXiv:2510.13197},
  year={2025}
}

@article{li2024nlp,
  title={Nlp-adbench: Nlp anomaly detection benchmark},
  author={Li, Yuangang and Li, Jiaqi and Xiao, Zhuo and Yang, Tiankai and Nian, Yi and Hu, Xiyang and Zhao, Yue},
  journal={arXiv preprint arXiv:2412.04784},
  year={2024}
}

@inproceedings{ruff2019self,
  title={Self-attentive, multi-context one-class classification for unsupervised anomaly detection on text},
  author={Ruff, Lukas and Zemlyanskiy, Yury and Vandermeulen, Robert and Schnake, Thomas and Kloft, Marius},
  booktitle={Proceedings of the 57th Annual Meeting of the Association for Computational Linguistics},
  pages={4061--4071},
  year={2019}
}

@inproceedings{manolache2021date,
  title={Date: Detecting anomalies in text via self-supervision of transformers},
  author={Manolache, Andrei and Brad, Florin and Burceanu, Elena},
  booktitle={Proceedings of the 2021 conference of the North American chapter of the association for computational linguistics: Human language technologies},
  pages={267--277},
  year={2021}
}

@inproceedings{das2023few,
  title={Few-shot anomaly detection in text with deviation learning},
  author={Das, Anindya Sundar and Ajay, Aravind and Saha, Sriparna and Bhuyan, Monowar},
  booktitle={International Conference on Neural Information Processing},
  pages={425--438},
  year={2023},
  organization={Springer}
}

@inproceedings{devlin2019bert,
  title={Bert: Pre-training of deep bidirectional transformers for language understanding},
  author={Devlin, Jacob and Chang, Ming-Wei and Lee, Kenton and Toutanova, Kristina},
  booktitle={Proceedings of the 2019 conference of the North American chapter of the association for computational linguistics: human language technologies, volume 1 (long and short papers)},
  pages={4171--4186},
  year={2019}
}

@article{liu2019roberta,
  title={Roberta: A robustly optimized bert pretraining approach},
  author={Liu, Yinhan and Ott, Myle and Goyal, Naman and Du, Jingfei and Joshi, Mandar and Chen, Danqi and Levy, Omer and Lewis, Mike and Zettlemoyer, Luke and Stoyanov, Veselin},
  journal={arXiv preprint arXiv:1907.11692},
  year={2019}
}

@inproceedings{kriegel2011interpreting,
  title={Interpreting and unifying outlier scores},
  author={Kriegel, Hans-Peter and Kroger, Peer and Schubert, Erich and Zimek, Arthur},
  booktitle={Proceedings of the 2011 SIAM International Conference on Data Mining},
  pages={13--24},
  year={2011},
  organization={SIAM}
}

@article{xu2024calibrated,
  title={Calibrated one-class classification for unsupervised time series anomaly detection},
  author={Xu, Hongzuo and Wang, Yijie and Jian, Songlei and Liao, Qing and Wang, Yongjun and Pang, Guansong},
  journal={IEEE Transactions on Knowledge and Data Engineering},
  volume={36},
  number={11},
  pages={5723--5736},
  year={2024},
  publisher={IEEE}
}

@inproceedings{qian2026dynhd,
  title={DynHD: Hallucination Detection for Diffusion Large Language Models via Denoising Dynamics Deviation Learning},
  author={Qian, Yanyu and Tan, Yue and Liu, Yixin and Yu, Wang and Pan, Shirui},
  booktitle={Findings of the Association for Computational Linguistics: EMNLP 2026},
  year={2026}
}

@inproceedings{tan2026influence,
  title={Influence-oriented personalized federated learning},
  author={Tan, Yue and Long, Guodong and Jiang, Jing and Zhang, Chengqi},
  booktitle={IEEE International Conference on Data Mining},
  year={2026}
}

@inproceedings{zhao2026fedcigar,
  title={FedCIGAR: A Personalized Reconstruction Approach for Federated Graph-level Anomaly Detection},
  author={Zhao, Yunfeng and Liu, Yixin and Chen, Qingfeng and Li, Shiyuan and Tan, Yue and Pan, Shirui},
  booktitle={IJCAI},
  year={2026}
}

@inproceedings{tan2023taming,
  title={Taming heterogeneity to deal with test-time shift in federated learning},
  author={Tan, Yue and Chen, Chen and Zhuang, Weiming and Dong, Xin and Lyu, Lingjuan and Long, Guodong},
  booktitle={International Workshop on Federated Learning for Distributed Data Mining},
  year={2023}
}

@inproceedings{chen2025multi,
  title={Multi-Stage Verification-Centric Framework for Mitigating Hallucination in Multi-Modal RAG},
  author={Chen, Baiyu and Wongso, Wilson and Hu, Xiaoqian and Tan, Yue and Salim, Flora D},
  booktitle={2025 KDD Cup Workshop for Multimodal Retrieval Augmented Generation},
  year={2025}
}

@article{li2026towards,
  title={Towards Anomaly Detection on Relational Data},
  author={Li, Shiyuan and Zhao, Yunfeng and Tan, Yue and Chen, Qingfeng and Liu, Yixin and Pan, Shirui},
  journal={arXiv preprint arXiv:2606.18621},
  year={2026}
}

@article{shen2026raising,
  title={Raising the bar in graph ood generalization: Invariant learning beyond explicit environment modeling},
  author={Shen, Xu and Liu, Yixin and Wang, Yili and Miao, Rui and Dai, Yiwei and Pan, Shirui and Chang, Yi and Wang, Xin},
  journal={IEEE Transactions on Pattern Analysis and Machine Intelligence},
  year={2026}
}

@inproceedings{li2026ofa,
  title={OFA-MAS: One-for-all multi-agent system topology design based on mixture-of-experts graph generative models},
  author={Li, Shiyuan and Liu, Yixin and Zheng, Yu and Li, Mei and Nguyen, Quoc Viet Hung and Pan, Shirui},
  booktitle={Proceedings of the ACM Web Conference 2026},
  pages={1333--1344},
  year={2026}
}

@article{liu2026few,
  title={From few-shot to zero-shot: Towards generalist graph anomaly detection},
  author={Liu, Yixin and Li, Shiyuan and Zheng, Yu and Chen, Qingfeng and Zhang, Chengqi and Yu, Philip S and Pan, Shirui},
  journal={IEEE Transactions on Knowledge and Data Engineering},
  year={2026},
  publisher={IEEE}
}

@inproceedings{liu2026rethinking,
  title={Rethinking Feature Alignment in Generalist Graph Anomaly Detection: A Relational Fingerprint-based Approach},
  author={Liu, Yujing and Liu, Yixin and Zheng, Yu and Liew, Alan Wee-Chung and Cao, Xiaofeng and Pan, Shirui},
  booktitle={International Conference on Machine Learning},
  year={2026}
}

@inproceedings{pan2026camera,
  title={CAMERA: Adapting to Semantic Camouflage in Unsupervised Text-Attributed Graph Fraud Detection},
  author={Pan, Junjun and Liu, Yixin and Zheng, Yu and Chi, Lianhua and Liew, Alan Wee-Chung and Pan, Shirui},
  booktitle={International Joint Conference on Artificial Intelligence},
  year={2026}
}

@article{zheng2026unsupervised,
  title={From unsupervised to few-shot graph anomaly detection: A multi-scale contrastive learning approach},
  author={Zheng, Yu and Jin, Ming and Liu, Yixin and Chi, Lianhua and Phan, Khoa T and Chen, Yi-Ping Phoebe},
  journal={Transactions on Graph Intelligence and Network Applications (TGINA)},
  year={2026}
}

@inproceedings{li2026towards2,
  title={Towards One-for-All Anomaly Detection for Tabular Data},
  author={Li, Shiyuan and Liu, Yixin and Zheng, Yu and Cao, Xiaofeng and Pan, Shirui and Shen, Heng Tao},
  booktitle={International Conference on Machine Learning},
  year={2026}
}

@inproceedings{liu2026beyond,
  title={Beyond a Single Perspective: Text Anomaly Detection with Multi-View Language Representations},
  author={Liu, Yixin and Yan, Kehan and Li, Shiyuan and Chen, Qingfeng and Pan, Shirui},
  booktitle={Joint European Conference on Machine Learning and Knowledge Discovery in Databases},
  year={2026}
}

\end{document}